# Change Detection between Multimodal Remote Sensing Data Using Siamese CNN

Zhenchao Zhang[1], George Vosselman[1], Markus Gerke[2], Devis Tuia[3], and Michael Ying Yang[1]

[1] University of Twente
[2] Technical University of Brunswick
[3] Wageningen University and Research
michael.yang@utwente.nl

**Abstract.** Detecting topographic changes in the urban environment has always been an important task for urban planning and monitoring. In practice, remote sensing data are often available in different modalities and at different time epochs. Change detection between multimodal data can be very challenging since the data show different characteristics. Given 3D laser scanning point clouds and 2D imagery from different epochs, this paper presents a framework to detect building and tree changes. First, the 2D and 3D data are transformed to image patches, respectively. A Siamese CNN is then employed to detect candidate changes between the two epochs. Finally, the candidate patch-based changes are grouped and verified as individual object changes. Experiments on the urban data show that 86.4% of patch pairs can be correctly classified by the model.

**Keywords:** Change detection; Siamese CNN; Laser scanning; Aerial images; Multimodal data

## 1 Introduction

Detecting topographic changes and keeping the topographic data up-to-date in urban scenes are important tasks in urban planning and environmental monitoring [33]. Nowadays, remote sensing data over urban scenes can be acquired through satellite or airborne imaging, Airborne Laser Scanning (ALS), Synthetic Aperture Radar (SAR), etc. In practice, remote sensing data are often obtained from different platforms and thus show different attributes. For this reason, change detection is challenging because the information volume and data characteristics between multimodal data differ strongly. This paper aims to detect building and tree changes between 3D airborne laser scanning data and multiview 2D aerial images (Fig. 1). Since national mapping agencies often possess large archives of aerial images and perform new data collection campaigns of both images and laser scanning data on a regular basis, our proposed method can be applied routinely and on several urban areas across the globe.

The need for a true multimodal method is dedicated by the fact that a direct comparison between 3D laser point clouds and 2D images is impossible. Change detection can be performed only after data transformation. One could transform the images into a point cloud, for instance through dense image matching [15,9,27]. However, the point clouds from laser scanning and from dense matching still differ in geometric accuracy, noise level, density and the amount of occlusion. Additionally, laser scanning data do not contain any spectral information.

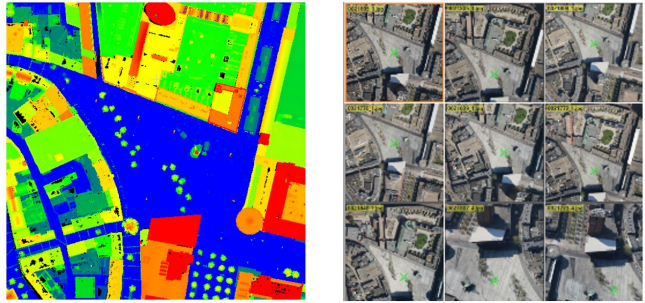

**Fig. 1.** Detect topographic changes between laser scanning data and aerial images. Left: laser scanning points (colored according to the height); Right: nine exemplary aerial images from multiple views over the same area.

Alternatively, the 3D point clouds can be transformed to 2.5D Digital Surface Models (DSMs) by meshing and interpolation techniques [19]. However, still a direct differencing between laser scanning-derived DSM and image-derived DSM, or the respective point cloud, reveals some data problems (Fig. 2): (1) Dense matching points on smooth terrain and roof surfaces usually contain much more noise than the laser scanning points; (2) Laser scanning can penetrate the tree canopy but dense matching cannot. As a consequence, laser scanning points are distributed over the canopy, branches and the ground below, while dense matching usually only delivers points on the canopy; (3) Dense matching leads to many false matches or noise in area with poor texture or heavy occlusion, e.g. along the narrow streets, in shadows or tree canopy; (4) Small horizontal registration errors between the two DSMs cause thin areas of false detection along some building edges; (5) Data gaps exist in both data sets [43], e.g. see the data gaps on the building roofs in the laser points, as well as the gaps along the alley in the dense matching points in the top row of Fig. 2. To cope with these problems, one would like to learn a transformation that reduces false change detections and highlights changed regions. Recently, Siamese CNN (S-CNN) architecture has shown superiority to traditional methods in comparing patches and detecting changes [39,24]. In this paper, we employ a fine-tuned S-CNN to detect patch-based topographic changes. First, the laser scanning-derived DSM and image-derived DSM are converted to gray image patches, respectively. The S-CNN is used to distinguish the true building and tree changes from the false changes caused by the data problems. The final output of the S-CNN is the change likelihood between the two inputs. Finally, the changed patches are grouped and verified into individual changed objects.

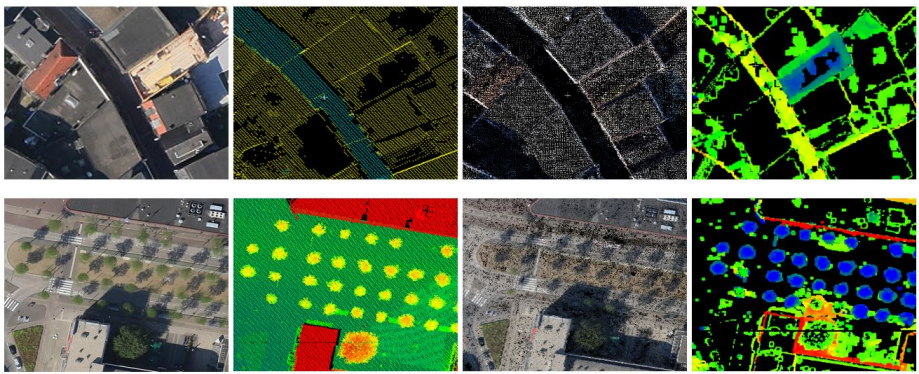

**Fig. 2.** Visualization of the problems in multimodal data. From left to right: ortho images for reference, laser scanning points colored by height, dense matching points with true scene colors, DSM differencing colored by height. The top row shows the scene of a narrow street; the bottom row shows the scene with tall and short trees.

To summarize, we make the following contributions:

- We present a change detection framework between 3D laser scanning points and 2D aerial images. The object heights are converted to gray image values using a non-linear function which enables a detailed representation of the object heights.
- An S-CNN architecture is employed to learn the different characteristics between multimodal remote sensing data and detect changes between different epochs. The S-CNN architecture is fine-tuned to distinguish the real building changes from false changes caused by mis-registration errors or false matches.
- A robust verification and grouping processing is proposed to detect individual object-level changes.
- We will make our code and dataset online available upon the acceptance of the paper.

This paper is structured as follows: related work is discussed in Sec. 2. The proposed framework is discussed in detail in Sec. 3. Experimental results of the proposed framework are shown and analyzed in Sec. 4. Finally, conclusion in Sec. 5 summarizes this paper.

## 2 Related work

Change detection is the process of defining differences in an object by analyzing it at different epochs [29]. The input data can be 2D images [22],3D point clouds [26], or 2.5D DSMs [32]. Change detection methods can be divided into two categories [41]: post-classification comparison [38,37,10,42,14,23] and change vector analysis [7,34]. In post-classification comparison, a pre-classification is required for both epochs. Change detection is then performed by comparing the labels between the classification maps obtained at each time step. When the data of two epochs are of different modalities, both the training and testing have to be performed on each data set, separately, which requires more time. Moreover, errors tend to be multiplied along object borders and give specific error in the single classification maps [35].

Change vector analysis allows a direct data comparison between two epochs. It relies on extracting handcrafted features based on prior knowledge of the scene and objects and fuses the change indicators in the final stage [25,5,32,8]. However, traditional change vector analysis is sensitive to the data problems and usually causes many false detections, especially in multimodal change detection. The key of change vector analysis is to extract representative feature vectors from the two epochs. Some previous work defines such transformations on eigenvector-based systems [34]. Recently, deep convolutional neural networks demonstrated their capabilities in extracting representative features for various tasks, e.g. image classification [17,30], semantic segmentation [21,28,1], object tracking [4], etc. As a specific architecture of CNN, S-CNN shows good performance in applications that require to compute similarities or detect changes between two input images. The application of S-CNN can be either patch-based (single value output) or dense (dense per-pixel output), depending on the expected output.

Bromley et al.[3] use a Siamese CNN to verify the handwritten digits and achieves an accuracy of 95.5%. A pair of image patches is passed through the two branches of the Siamese networks to generate a pair of feature vectors. The similarity score of the features vectors is compared to the training label. The gradient of the loss function with respect to the network parameters is computed using back-propagation. The S-CNN allows end-to-end training and no handcrafted feature extraction is required. The results show that the Siamese CNN is robust to large intra-class variations. Patch-based comparison can also be found in other applications, such as face verification [6,31], character recognition [16] and patch matching [13]. In the remote sensing domain, S-CNN has been used to identify corresponding patches between SAR images and optical images [24]. Although SAR images and aerial images show heterogeneous properties, they achieve an overall accuracy of 97.48%. [20] matches street-view images to the city-scale aerial view images so as to geo-localize the street images. The image patches contain large variations in illumination, rotation and scale. Similarly, [18] uses S-CNN to perform change detection between aerial images and street level panoramas reprojected on an aerial perspective.

In dense prediction applications, [39,40] use a Siamese CNN for wide-baseline stereo matching. The S-CNN takes a pair of image patches as input and outputs a similarity score. The similarity score for each pixel is aggregated into a global matching cost. Minimizing the matching cost will lead to the final disparity maps. [41] uses an S-CNN to detect changes in pairs of optical aerial images at pixel level. Each convolutional layer maintains the original input size, so that a dense change inference map is obtained.

## 3 Proposed framework

### 3.1 Pre-processing

The 2D aerial images and 3D laser scanning point clouds should be transformed to 2D image patches before they are fed into the S-CNN. To do so, we use the following procedure:

- Images (2D) to point cloud (3D): the Patch-based Multi-View Stereo method (PMVS) is used to generate dense point clouds from images [9].
- Point cloud (3D) to DSM (2.5D): the point clouds are then transformed to 2.5D DSM by assigning the height of the highest point within a cell as the cell value. If there is no point within a cell, this cell is marked *empty*.
- DSM (2.5D) to nDSM (2.5D): the normalized DSM is then obtained by subtracting the Z value of a Digital Terrain Model (DTM) from the DSM at every grid cell. The DTM is obtained by filtering the laser scanning data using progressive TIN densification [2]. The same DTM is used to filter point clouds from laser scanning and dense matching, respectively. Taking the nDSM as input to a S-CNN is more informative than the original DSM, because the absolute height in the nDSM corresponds to the object height, which is usually useful for scene parsing. As a comparison, geodetic heights are recorded in the DSMs. The absolute values in the DSMs contain less information about the objects, especially if the terrain is sloped.
- nDSM (2.5D) to gray images (2D): the nDSM is finally converted to gray images. The height at each DSM cell is converted to an integral gray value, which ranges in [0, 255]. The transformation function is adapted from the inversely proportional function using Eq.(1):

$$v = \begin{cases} \dfrac{255(x-t)}{(x-t)+q} & \text{if } x \geq t, \\ 0 & \text{if } x < t. \end{cases} \quad (1)$$

where $q$ is the parameter controlling the steepness of the curve and $t$ is to control the translation of the curve on $X$ axis. In some areas, the terrain height might be negative, for instance if there is construction work on the site. To cope with these cases, we introduce a truncation threshold $t$. If the height within one cell is even lower than $t$, the gray value is set to 0.

Heights larger than t are scaled to [0, 255] using Eq. 1 with parameters $q = 10$ and $t = -2$, providing the curve in Fig. 3. The curve is quite steep from [-2 m, 40 m], which is a common height range for objects in real scenes. The object heights in this range will be converted to gray values from 0 to 200 with little loss of resolution. When the height is larger than 40 m, the gray value will further go up and get closer to 255. We did not use a direct linear transformation for two reasons: (1) the linear transformation cannot represent a height value greater than a threshold; (2) the linear transformation makes equal scaling to all the height values. However, more details exist for low objects with a height lower than 40 m in urban scene.

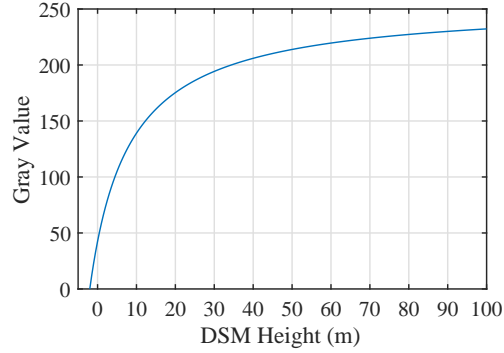

**Fig. 3.** Curve with $q = 10$ and $t = -2$ for converting DSM heights to gray values.

After converting the 2D images, the laser scanning points are also converted to 2D gray value images following a similar workflow with the same transformation parameters. After data conversion, the 2D gray images from the two sources (laser scanning and dense matching) are cropped into square patches of the same size (see Fig. 4). Two patches belonging to a pair represent exactly the same location in the study area.

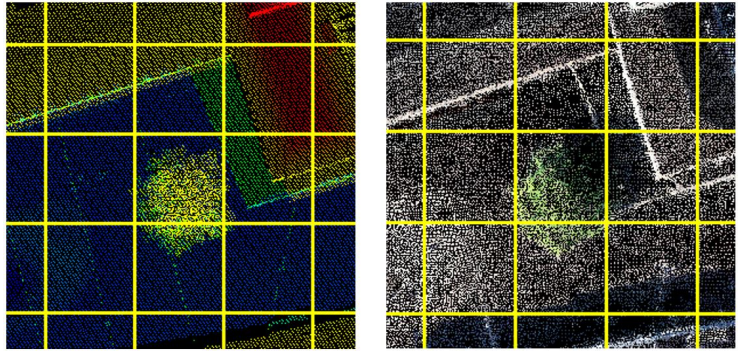

**Fig. 4.** Patch-based change detection. Left: laser scanning data; right: dense matching data.

Concerning data gaps in the patches, since data gaps on the two patches are not at the same locations and differ in quantities, the data gaps will also lead to some false differences between the two data sets. Given a pair of patches $X$ and $Y$, a conjugated processing for the patch pair includes two steps: (1) If data gap occurs in a certain pixel on the image $X$, then set the same location on the image $Y$ to *data gap*. Do this in bi-directions. (2) If the data gap ratio in the image patch is higher than 50%, this pair is regarded as invalid and not considered in training or testing. After processing, the presence of data gaps will not influence the change detection.

### 3.2   Network architecture

A Siamese CNN architecture is used to detect candidate building and tree changes between the two gray value patches. The Siamese CNN is a variant of a traditional CNN, which includes two branches for the input and concatenates the outputs of each branch into a new features map or feature vector in the end [39]. Our S-CNN architecture is adapted from the code originally written for face verification [11]. Since there is no rotational variance or scale variance between the patch pairs in our case, we use an S-CNN architecture with fewer layers compared to [24,41]. The architecture and detailed configurations are shown in Fig. 5. In the S-CNN architecture, each CNN branch includes three convolutional and three fully connected blocks. The output of each CNN branch is a 5-dimensional vector.

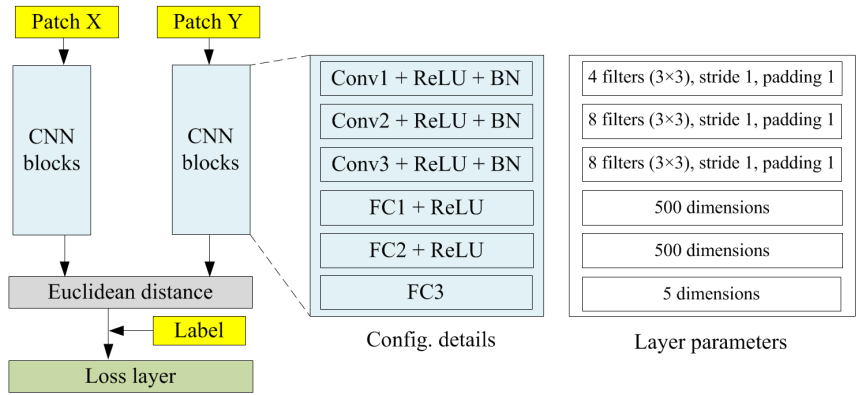

**Fig. 5.** The S-CNN architecture used for change detection. Left: architecture of the Siamese CNN; middle: the feature extraction network with configuration details; right: parameters of each layer.

The two branches may share weights or not depending on how much the two inputs are similar to each other [39]. We used S-CNN with shared weights in our configuration for two reasons: (1) after pre-processing the two DSMs are mainly similar in the unchanged areas; (2) we have only limited training samples. A S-CNN with more layers requires more data to train; Considering the number of parameters in the twin branches, S-CNN with shared weights contains half the parameters compared to a S-CNN with unshared weights.

During training, the input to the S-CNN is a pair of image patches $X$ and $Y$ derived from laser scanning points and aerial images, respectively. The twin CNN branches work as a feature extractor. Let the CNN branch be represented by $f(.)$. Then the outputs of each branch are $f(X)$ and $f(Y)$, respectively. The Euclidean distance $D$ (cf. Eq.(2)) is the distance metric to measure the similarity between $X$ and $Y$ in the feature space [12,20]. The larger $D$ is, the more likely a change has occurred.

$$D = ||f(X) - f(Y)||_2 \qquad (2)$$

During optimization, we minimize the loss function through maximizing $D$ between changed pairs and minimizing $D$ between unchanged pairs. The contrastive loss for a pair of patches proposed by [12] is used:

$$E(X, Y, l_T) = (1 - l_T) D^2 + l_T \left(\max\left(0, m - D\right)\right)^2 \qquad (3)$$

where $l_T \in \{0, 1\}$ is the true label which equals 1 for changed pairs and 0 for unchanged pairs. $m$ is to control the margin between changed and unchanged pairs. Minimizing the sum of the loss in one iteration penalizes unchanged patches by $D^2$, and penalizes changed patches by $(m - D)^2$, where $|m - D|$ is the distance to the margin for distances smaller than $m$.

During inference, we compute $f(X)$ and $f(Y)$ in the feed-forward network and then compute the Euclidean distance. The Euclidean distance is an indicator for how likely the patches are changed. Meanwhile, we can judge whether the inference is correct. For instance, we can use 0.5 as hard threshold to distinguish changed and unchanged pairs. Supposing $l_P$ is the predicted Euclidean distance ($l_P \geq 0$), then the prediction is correct only if $|l_P - l_T| < 0.5$.

### 3.3 Detecting individual object changes

The outcome of S-CNN informs about which patches have undergone a change. Since we are aiming for individual object-based changes, the neighboring changed patches need to be grouped into individual objects.

**Group changed patches into individual objects.** This grouping process contains three steps (see Fig. 6):

Step 1: The changed patches are connected and expanded outwards for half patch size in order to contain the complete changed objects (Fig.6(b)). The point cloud of the candidate changed area is filtered using progressive TIN densification (Fig.6(c)).

Step 2: The non-ground points are then segmented using a surface-based growing algorithm to extract planes. In surface growing, the initial planes are first detected using 3D Hough transform. Then a surface growing radius of 1.0 m and maximum distance between point and fitted plane of 0.2 m are employed according to the point cloud density and the noise level [36]. After surface growing, mainly the building points are segmented to planes. The remaining clutter points, which do not belong to any plane, are further grouped into complete clutters using connected component analysis. The result of this segmentation is illustrated in Fig. 6(d). The building planes are merged into building segments and tree points are merged into tree clutters for each individual tree.

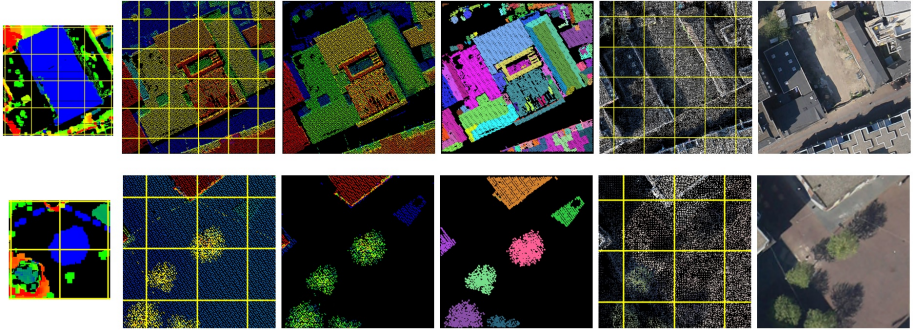

**Fig. 6.** Workflow for grouping individual changed objects. Top row: a true building change; Bottom row: a false tree change. From left to right: (a) changed patches on the height differencing maps, (b) expanding the boundary of the laser scanning data outward by half patch size as the new Region Of Interest (ROI), (c) point cloud filtering, (d) point cloud segmentation, (e) dense matching points for reference, (f) orthoimages for reference. The yellow lines indicate the patch boundary.

Step 3: Segment-based classification: After the individual objects in the candidate area are separated, segment-based features can be calculated for each segments to classify them into either building or tree using some knowledge-based rules. The segment-based features include *segment size*, *planarity of segment*, *plane slope*, *average angle* and *residual of plane fitting* [43].

**Change detection and verification.** After classification, the segments are classified into three classes: *roof segment*, *tree clutter* and *other*. In the cases of the two first, change detection at the object level is performed as follows:

– If a segment is a *roof segment*, it is compared to the neighboring points in the other point set to check the deviation between them. If the plane to the other point cloud is smaller than a certain threshold (e.g. 0.5 m), this building plane is labeled as *unchanged*; otherwise it is labeled as *changed*.
– If a segment is a *tree clutter*, in this case the distance to the dense matching point cloud can not indicate whether it is changed or not. The reason is that dense matching sometimes fails on the tree canopy with sparse twigs. Therefore, the canopy changes between the two point clouds cannot indicate whether the tree is changed in the real scene. In this case, the tree change can be verified in the original aerial images and orthoimages using the *normalized Excessive Green Index* (*nEGI*) [26]:

$$nEGI = \frac{2G - R - B}{2G + R + B} \qquad (4)$$

If the *nEGI* values on the relevant image pixels are larger than a certain threshold, this area is labeled vegetation; Namely, the vegetation is still there on the images. Therefore, there is no change in the pair. Since a 3D point is visible on multiple oblique aerial images, the verification can be performed on different images. The 3D tree points are projected back to the aerial images using:

$$(x, y, 1)^T = P(X_0, Y_0, Z_0, 1)^T \qquad (5)$$

where (x, y) are the 2D coordinates on aerial images and $(X_0, Y_0, Z_0)$ are the world coordinates in the 3D space. $P$ is a $3 \times 4$ matrix that combines the intrinsic and extrinsic parameters:

$$P = K_{int} \begin{bmatrix} R & -RT \\ 0 & 1 \end{bmatrix} \qquad (6)$$

where $K_{int}$ is the intrinsic parameter related to the camera calibration, $R$ is the rotation matrix and $T$ is the translation matrix.

Grouping an individual object can be performed also on the dense matching points. If a new object appears in the new dense matching point cloud, the same filtering and classification can be used. The roof segments and the tree clutters can then be verified with the laser scanning points. Note that no matter whether a tree appears or disappears in the dense matching points, the change decision should always be verified on the aerial images and orthoimages.

## 4 Experiments

### 4.1 Training Setup and Implementation Details

The laser scanning points and oblique aerial images were acquired over Enschede, the Netherlands in 2007 and 2011, respectively. The study area is 1.6 $km^2$ wide as shown in Fig. 7. It is divided into 9 tiles. Tile 4, 5, 7, 8 are used for training. Tile 1 is used for validation. Tile 2, 3, 6, 9 are used for testing.

Square patches are cropped from the point clouds without overlapping. The patch size is $10\,m \times 10\,m$. The patch size should be large enough to contain sufficient object content yet small enough to ensure a detailed change detection. It is common that a single building or tree may span over more than one patch. The pairs are labeled based on the comparison between point clouds from laser scanning and dense matching with the guidance of orthoimages and DSM differencing maps. Three rules are considered when labeling the pairs: (1) When a building is partly changed (either new or demolished) between the patches, this pair is labeled as *changed*; (2) If a tree is successfully matched in the dense matching points, it is labeled as *unchanged*; if the tree is not matched, this patch is labeled as *changed*. Considering the tree changes, we are only interested in whether a tree appears or disappears between the two point clouds; in other words, we do not detect the natural tree growth. (3) We neglect changes due to cars, data noise, subtle point cloud mis-registration, etc.

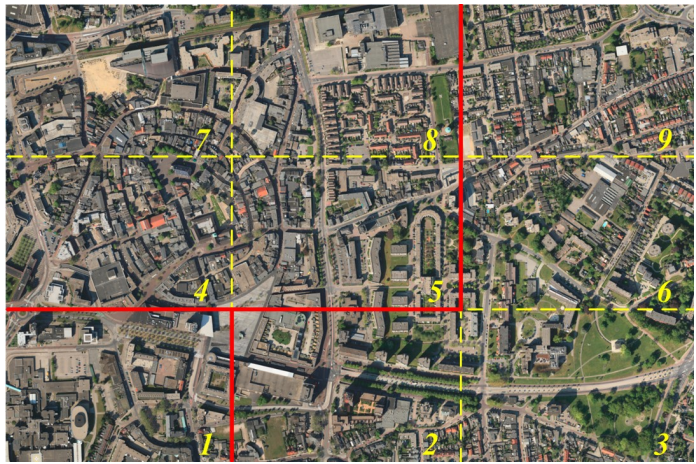

**Fig. 7.** The orthoimage of the study area. The study area is divided into 9 tiles.

Each pair of point cloud patches is converted to a pair of gray image patches using the procedure detailed in Section 3.1. Some examples (both changed and unchanged) are shown in Fig. 8. The size of gray image patch is $100 \times 100$. In this case, one pixel in the gray image is equivalent to 0.1 m in the urban scene, which is relatively detailed representation of the object surface. Considering the size of the training data, there are fewer changed pairs than unchanged pairs. In order to generate comparable number of changed and unchanged samples, data augmentation is performed on the changed training pairs. Each pair is horizontally and vertically flipped and also rotated by 90°, 180°, and 270° [41] and the same transformation is applied to both images of the pair. With this procedure, we obtain 5058 unchanged pairs and 7458 changed pairs for training. 800 pairs are used for validation. 6319 pairs are used for testing which contain 2165 changed pairs and 4154 unchanged pairs.

In the training process, we train the network from scratch. The batch size is set to 128. Changed and unchanged pairs are randomly selected for each batch. The learning rate is set to 0.0001. The training is run for 10 epochs. The training takes approximately 25 minutes on a single NVIDIA Titan X GPU. The training loss and validation accuracy are shown in Fig. 9. *Accuracy* indicates the ratio of right predictions (both changed and unchanged) to the total number of tested pairs. After around 600 iterations, the training loss tends to stabilize and the validation accuracy reaches 88.4%.

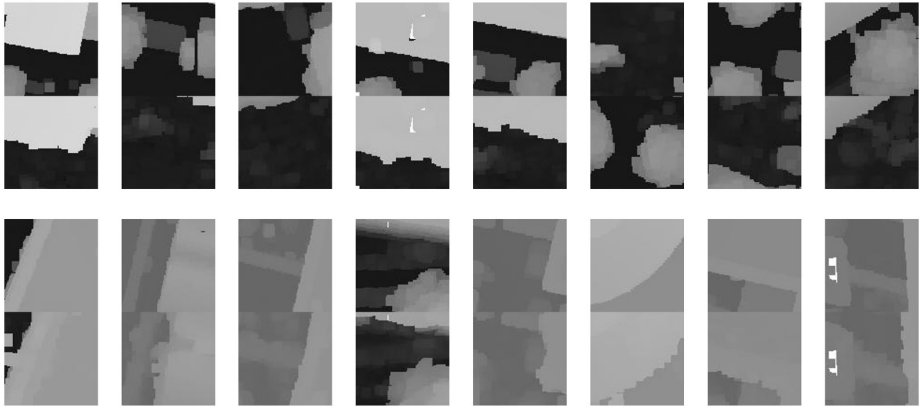

**Fig. 8.** Changed and unchanged pairs for training. Top: changed pairs; bottom: unchanged pairs. Row 1 and row 3 are converted from laser scanning points, while row 2 and row 4 are from aerial images.

### 4.2   Testing Results

**Evaluation metrics.** Both precision and recall are calculated based on patches: $precision = \frac{TP}{TP+FP}$ and $recall = \frac{TP}{TP+FN}$. *True Positive* ($TP$) is the number of correctly detected changes. *True Negative* ($TN$) is the number of correctly detected unchanged pairs. *False Positive* ($FP$) is the number of unchanged pairs predicted as changed. *False Negative* ($FN$) is the number of changed pairs predicted as unchanged.

The testing results are as follows: *TP* is 1746, *TN* is 3715, *FN* is 419, *FP* is 439. The testing accuracy is 86.4%, which is slightly worse than the validation accuracy. Considering the change detection, the precision is 79.9% and the recall is 80.6%. Some testing results are visualized in Fig. 10. The trained S-CNN is able to learn the change patterns of new buildings, demolished buildings, new trees and removed trees. It also recognizes some false changes between the two point clouds caused by tree growth, dense matching errors, or mis-registration errors.

Figure 10(c) illustrates some falsely detected changes. Some patches with car changes are detected as change. Some patches with large dense matching errors are incorrectly marked as changed as well (see the pairs with Euclidean distances of 0.525 or 0.567, respectively). In Fig. 10(d), some small areas of tree changes are missed by the S-CNN model which leads to some omission errors.

Finally, the patch-based changes are grouped and verified into individual changed objects. The final object-based changes are shown in Fig. 11. All the changed buildings can be correctly detected (Fig. 11(a-c)). Considering the tree changes, all the newly-planted trees in the dense matching points can be correctly detected (Fig. 11(d-e)). There are many trees labeled as *disappeared* between the two point clouds by the S-CNN. Actually, many trees disappear in the dense matching points because dense matching fails on the tree canopy with sparse twigs (see Fig. 11(f)). In the following verification step using raw images and orthoimages, most of them are verified to be unchanged trees.

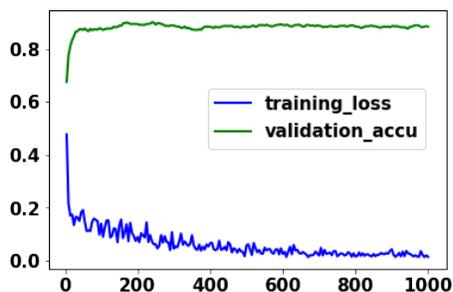

**Fig. 9.** Training loss and validation accuracy. The X-axis shows the number of iterations.

Comparing with other multimodal change detection methods [32,8], the proposed method does not need to extract handcrafted features or fuse the change decisions in the end. Considering efficiency, since inference of the S-CNN is very fast, the bottleneck mainly lies in data transformation and change verification. However, transformation to 2.5D and handling 2D gray images are not computationally expensive; change verification is only required in some candidate areas, which is computationally efficient.

## 5  Conclusions

This paper proposes a method to detect building and tree changes between 3D laser scanning data and 2D aerial images. Dealing with the characteristics of the different sources of data is the main research challenge. Our solution is to first convert the point clouds and 2D imagery to corresponding 2D gray image patches using photogrammetric methods. The patch pairs are then fed into a Siamese CNN with the Euclidean distance of the feature vectors as the output. The S-CNN model not only evaluates whether the changed patches are correctly inferred, but also gives an inference on how likely a certain pair is changed. The patch-based changes are further grouped and verified into individual object changes. Dense matching is often problematic on the tree canopy or in the shadow. The final step helps in verifying the detected changes on the aerial images and omitting false detections. Results on the urban data from the Netherlands demonstrate the effectiveness of the method, where the S-CNN model was able to distinguish real building and tree changes from false changes caused by the multimodal data differences. Future work will focus on designing the network architecture towards end-to-end learning and dense prediction of the changed locations.

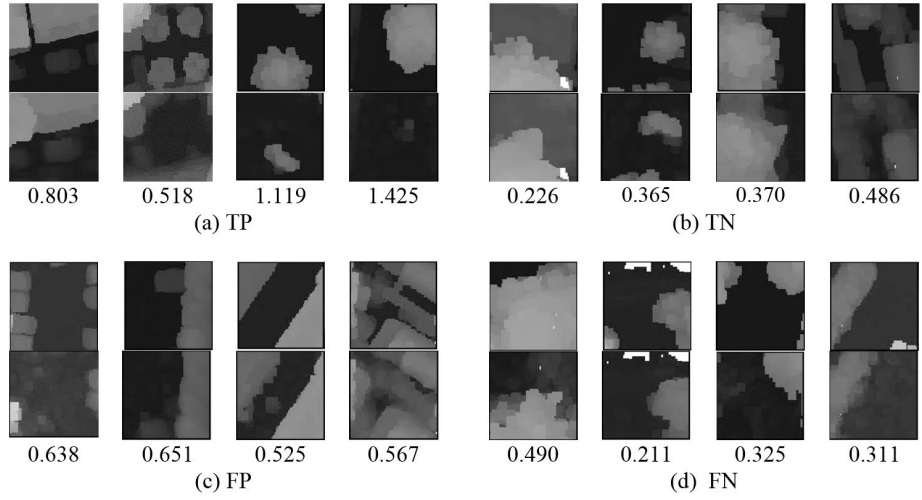

**Fig. 10.** Testing results with four examples for each classification outcome. In each group, the first row is the laser scanning patch, the second row is dense matching patch. The digits below each pair are the Euclidean distances computed by the S-CNN model.

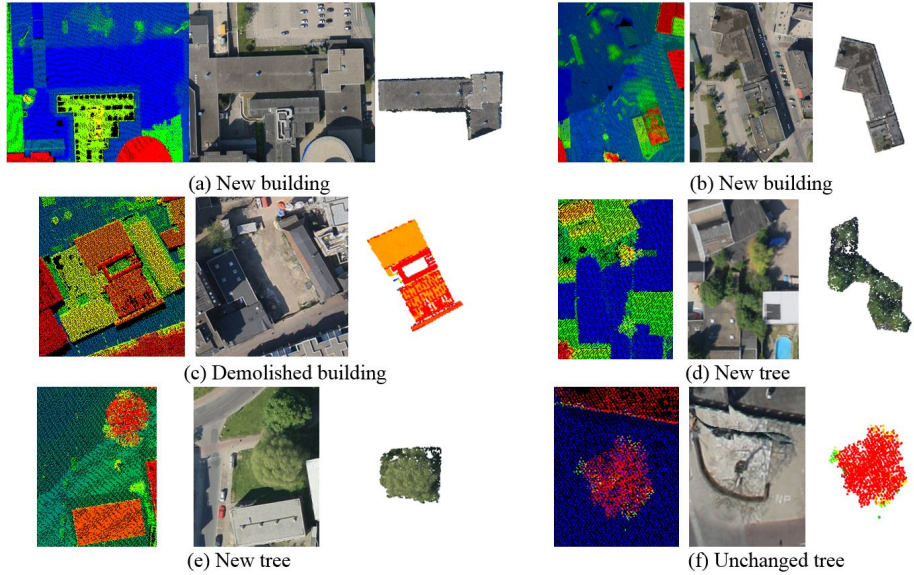

**Fig. 11.** Some detected object-level changes. Left: laser scanning points; middle: orthoimage; right: changed object detected by the S-CNN model.